\documentclass[lettersize,journal]{IEEEtran}

\usepackage{algorithm}
\usepackage{algorithmic}
\usepackage{amsfonts}
\usepackage{amsmath}
\usepackage{amssymb}
\usepackage{array}
\usepackage{booktabs}
\usepackage{colortbl}
\usepackage{caption}

\usepackage{enumitem}
\usepackage{fontawesome5}
\usepackage{graphicx}
\usepackage{hyperref}
\usepackage{listings}
\usepackage{longtable}
\usepackage{makecell}
\usepackage{mdframed}
\usepackage{multirow}
\usepackage{multicol}
\usepackage{newfloat}
\usepackage{pifont}
\usepackage{subcaption}
\usepackage[most]{tcolorbox}
\usepackage{tikz}
\usepackage[table]{xcolor} 
\usepackage{tabularx}
\usepackage{xspace}

\definecolor{lightgray2}{gray}{0.85}

\definecolor{myred}{RGB}{237,28,80 }
\definecolor{scarlet}{RGB}{255,36,0}
\definecolor{keywordred}{RGB}{200,34,34}
\definecolor{iceblue}{RGB}{214, 230, 245}
\definecolor{pastelyellow}{RGB}{254, 240, 158}
\definecolor{creamyellow}{RGB}{255,246,213}

\newcommand{\ourmethod}{StyleDecipher\xspace}
\newcommand{\ourmethodTT}{\textit{\textbf{\fontfamily{lmtt}\selectfont \ourmethod}}\xspace}

\newcommand{\GPTThreeFiveTurbo}{GPT-3.5-Turbo\xspace}

\begin{document}

\title{\ourmethod: Robust and Explainable Detection of LLM-Generated Texts with Stylistic Analysis}
\author{Siyuan Li, \IEEEmembership{\large Graduate Student Member, IEEE}, Aodu Wulianghai, Xi Lin, \IEEEmembership{\large Member, IEEE}, Guangyan Li, \\ Xiang Chen, Jun Wu, \IEEEmembership{\large Senior Member, IEEE}, Jianhua Li, \IEEEmembership{\large Senior Member, IEEE}
    \thanks{Siyuan Li, Xi Lin, Jun Wu, and Jianhua Li are with the School of Computer Science, Shanghai Jiao Tong University, Shanghai, China, and also with Shanghai Key Laboratory of Integrated Administration Technologies for Information Security, Shanghai, China 
    (Email: \{siyuanli, linxi234, junwuhn, lijh888\}@sjtu.edu.cn).}
    \thanks{Aodu Wulianghai is with the School of Computer Science, Shanghai Jiao Tong University, Shanghai, China (Email: \{melusine.wlhad\}@sjtu.edu.cn).}
    \thanks{Guangyan Li is with the State Key Laboratory of Multimodal Artificial Intelligence Systems, Institute of Automation, Chinese Academy of Sciences, Beijing 100190, China (Email: \{liguangyan2022\}@ia.ac.cn).}
    \thanks{Xiang Chen is with the College of Computer Science and Technology, Zhejiang University, Hangzhou 310007, China (Email: wasdnsxchen@gmail.com).}
}

\markboth{Submitted to IEEE Transactions on Dependable and Secure Computing}
{Shell \MakeLowercase{\textit{et al.}}: A Sample Article Using IEEEtran.cls for IEEE Journals}

\IEEEpubid{0000--0000/00\$00.00~\copyright~2025 IEEE}

\maketitle

\begin{abstract}
With the increasing integration of large language models (LLMs) into open-domain writing, detecting machine-generated text has become a critical task for ensuring content authenticity and trust. 
Existing approaches rely on statistical discrepancies or model-specific heuristics to distinguish between LLM-generated and human-written text.
However, these methods struggle in real-world scenarios due to limited generalization, vulnerability to paraphrasing, and lack of explainability, particularly when facing stylistic diversity or hybrid human-AI authorship. 
In this work, we propose \ourmethodTT, a robust and explainable detection framework that revisits LLM-generated text detection using combined feature extractors to quantify stylistic differences.
By jointly modeling discrete stylistic indicators and continuous stylistic representations derived from semantic embeddings, \ourmethod captures distinctive style-level divergences between human and LLM outputs within a unified representation space. 
This framework enables accurate, explainable, and domain-agnostic detection without requiring access to model internals or labeled segments. 
Extensive experiments across five diverse domains, including news, code, essays, reviews, and academic abstracts, demonstrate that \ourmethod consistently achieves state-of-the-art in-domain accuracy. 
Moreover, in cross-domain evaluations, it surpasses existing baselines by up to 36.30\%, while maintaining robustness against adversarial perturbations and mixed human-AI content. 
Further qualitative and quantitative analysis confirms that stylistic signals provide explainable evidence for distinguishing machine-generated text. 
Our source code can be accessed at \url{https://github.com/SiyuanLi00/StyleDecipher.}
\end{abstract}


\section{Introduction}
\IEEEPARstart{T}{he} increasing sophistication of large language models (LLMs) has enabled impressive advances in text generation across a variety of domains, including news, creative writing, and academic research~\cite{bubeck2023sparks, huang2024position}.
These models have been extensively deployed across various domains, fundamentally transforming text generation and interaction by producing content that exhibits a high degree of resemblance to human writing in both stylistic nuances and linguistic complexity.
These models have been extensively deployed across various domains, fundamentally transforming text generation and interaction. 
Content produced by them exhibits a high degree of resemblance to human writing in both stylistic nuances and linguistic complexity.
However, the widespread adoption of these models has raised significant concerns regarding the authorship attribution of content, especially in contexts where the boundary between human and machine authorship becomes ambiguous~\cite{wu2024DetectRL, yang2024survey, sadasivan2025can, wu2025survey, yu2025evobench}.
Distinguishing between human and machine authorship has become more difficult due to the generation of increasingly sophisticated texts. This complication affects tasks such as ensuring proper attribution, verifying the source of information, and maintaining trust in content creation~\cite{li2025prdetect}.
Detecting LLM-generated text has therefore emerged as a critical task, particularly in tackling toxic content~\cite{toxic:he2024you}, preventing misinformation~\cite{misinformation:zeng2024combining}, and safeguarding the authenticity of human news~\cite{news:wang2024explainable}.
\IEEEpubidadjcol

Traditional approaches to machine-generated text detection have primarily focused on leveraging statistical measures like perplexity, probability curvature, and various heuristic features to distinguish between human and LLM-generated texts~\cite{GLTR:gehrmann2019gltr, hashimoto2019unifying}. 
These early methods made significant strides by quantifying the likelihood of a text being generated by an LLM, relying on the assumption that the generative process of LLMs differs significantly from human writing in its statistical properties. 
While these methods have achieved notable success in controlled environments, such as testing on datasets generated with a fixed set of models or under specific conditions, they often struggle to generalize to more complex, real-world scenarios~\cite{hu2024radar, chen2025online}. 
For instance, many detection systems are heavily reliant on the output of a specific model or generation method, which limits their effectiveness when confronted with newer models or hybrid human-AI content~\cite{li2025model, cheng2025beyond, zhou2025adadetectgpt}, where the style of machine-generated text might be heavily influenced by human input or text revisions. 

As a result, these systems fail to account for stylistic variability across different models and the hybrid content generation, in which both humans and LLMs contribute to the final text~\cite{lei2025pald, StyleRepresentations:soto2024few}. 
This makes the detection task particularly challenging in dynamic, diverse environments. 
Furthermore, these approaches typically do not provide insight into the localized features that make certain sections of a text more likely to be generated by a machine, limiting their explainability and practical applicability~\cite{baseline2:Binoculars:hans2024spotting, chen2025online, bao2025glimpse}. 
This lack of explainability is a significant barrier to applying these systems in real-world settings, where understanding why a system classifies text as machine-generated is just as important as the classification itself~\cite{koike2025exagpt, teja2025fine}. 
Without a deeper understanding of the specific features contributing to a decision, these methods are less useful in complex, real-world applications where transparency and accountability are crucial. 
\begin{figure}[!t]
    \centering
    \includegraphics[width=\linewidth]{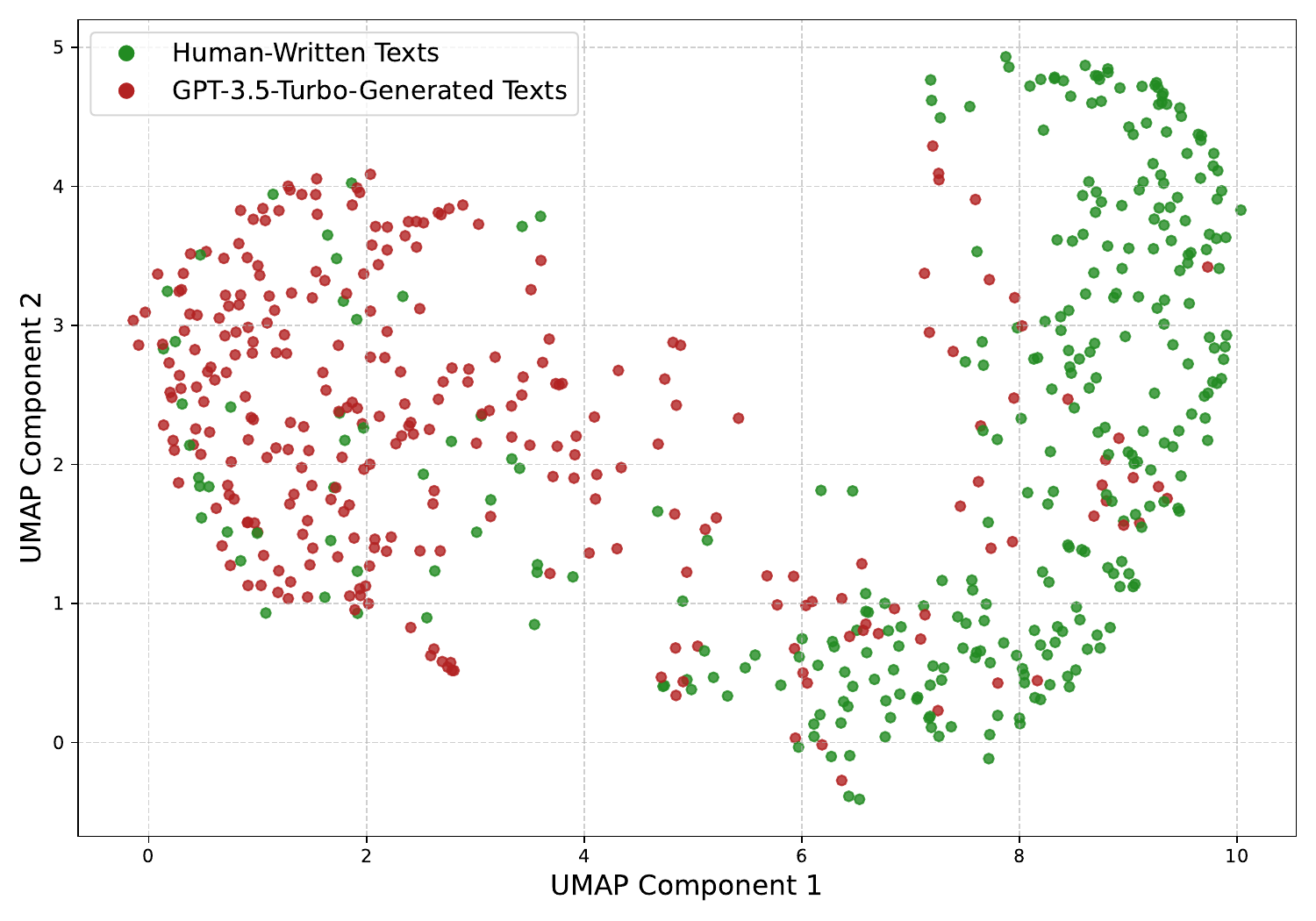}
    \caption{2D UMAP projection of \textit{\textcolor{keywordred}{\textbf{stylistic features for LLM-generated texts and human-written texts}}} on the Yelp Review dataset. 
    Each point represents an embedded paragraph ($\leq$ 64 tokens) encoded with combined stylistic features.
    \textit{\textbf{LLM-generated texts exhibit consistent stylistic patterns, forming distinguishable clusters from human-written texts}}.
    }
    \label{figure:scatter-plot}
\end{figure}

To address these limitations and enhance the utility of detection methods in real-world scenarios, recent advancements have explored increasingly sophisticated frameworks that incorporate style representations~\cite{StyleRepresentations:soto2024few} and perturbation robustness mechanisms~\cite{baseline3:RAIDAR:mao2024detecting} to improve detection accuracy and robustness. 
These approaches have shown promise in addressing some of the limitations of earlier methods, especially in controlled environments. 

However, current techniques remain limited in several crucial aspects. 
Most existing approaches depend heavily on model- or dataset-specific cues, which limits generalization to unseen tasks or new LLMs.
Further, explainability is often superficial: many detectors deliver a verdict without revealing which parts of the text evidence machine generation, nor why~\cite{bao2025glimpse, ji2024detecting}. 
This lack of transparency not only hinders their practical application in real-world scenarios but also reduces their effectiveness in settings where understanding the reasoning behind classification is critical.
Another under-served scenario is hybrid or mixed texts, in which human-written and LLM-generated segments coexist; existing detectors struggle to reliably detect such segment-level mixtures~\cite{zhang2024llm, lei2025pald}. 

Together, these deficiencies expose a pressing research gap: how can we build detectors that are robust, explainable, and capable of fine-grained detection, handling both pure and hybrid texts, while generalizing well across domains and models?
To address this gap, we focus on stylistic divergence as the central lens, integrating perturbation analysis to assess how stable or volatile style features are under semantic-preserving rewrites.
We frame our work around the following research questions:
\begin{itemize}
    \item \textbf{\textit{RQ1}: Robust Detection via Stylistic Divergence.} How can we quantify style stability under controlled rewriting to enhance detection robustness and domain/model generalization?
    \item \textbf{\textit{RQ2}: Explainable and Hybrid Attribution.} How can a detector not only decide if a text is machine-generated, but also which segments show stylistic divergence, with explainable evidence?
\end{itemize}

To address \textbf{RQ1}, we introduce \ourmethod, a \textit{stylistic divergence}–based detection framework, which redefines LLM-generated text detection as the task of measuring the stability of stylistic representations under controlled rewriting.
Instead of relying solely on lexical or statistical cues, our method encodes each text into a set of discrete and continuous style descriptors.
By applying a semantic-preserving rewriting model, we observe how these stylistic features change, quantifying stylistic stability as an indicator of authorship consistency.
This perspective enables our detector to achieve robustness and domain generalization, as the decision boundary is defined by fundamental stylistic behavior rather than dataset-specific signals.
To address \textbf{RQ2}, we extend this framework with a localized attribution mechanism that decomposes the overall stylistic divergence score into fine-grained hybrid contributions.
Instead of treating an entire document as a single decision unit, the model performs modular scoring on consecutive segments, measuring how local stylistic deviations align with machine-generation patterns.
This design enables explainable detection and highlights which spans contribute most to the classification outcome, offering transparent, actionable insights into the stylistic rationale behind its predictions.

Our main contributions are summarized as follows:
\begin{itemize}
    \item \textbf{\textit{Comprehensive stylistic divergence modeling}}: We introduce a framework that fuses discrete (n-gram overlap and edit distance) and continuous (semantic embedding) style features, evaluated under perturbations, thereby improving robustness and generalization.
    \item \textbf{\textit{Explainable and modular attribution framework}}: We develop a modular scoring mechanism for hybrid detection that does not require explicit span annotations, enabling identifiable evidence of stylistic divergence in hybrid human-AI texts.
    \item \textbf{\textit{Strong empirical validation on various datasets and settings}}: Extensive experiments across multiple benchmark datasets, including in-domain, out-of-domain, adversarial, and hybrid settings, demonstrate that \ourmethod outperforms existing baselines in detection accuracy, robustness, and explainability.
\end{itemize}

\section{Related Works}
In this section, we review previous work on detecting machine-generated text: (i) traditional machine-generated text detectors, (ii) watermark-based methods, (iii) statistical feature and supervised training-based methods, and (iv) more advanced methods tailored for real-world applications.

\subsection{Traditional Machine-Generated Text Detection}
Traditional methods for detecting machine-generated text rely on statistical analysis to identify anomalies between machine-generated and human-written content, using techniques such as \textit{perplexity, relative entropy, and log-likelihood}. 
Pioneering work by Lavergne et al.~\cite{lavergne2008detecting} applied relative entropy to identify discrepancies between the two types of text. 
Similarly, Hashimoto et al.~\cite{hashimoto2019unifying} showed that perplexity quantifies model uncertainty, providing a reliable method for distinguishing machine-generated content. 
Gehrmann et al.~\cite{GLTR:gehrmann2019gltr} introduced GLTR, which leverages features like per-word probability, rank, and entropy to detect textual anomalies. 
Although these methods laid the foundation for detection, they struggle with the complexity of modern LLMs, failing to capture the subtlety of contemporary AI-generated texts~\cite{jawahar2020automatic}.

\subsection{LLM-Generated Text Detection}
\subsubsection{Watermarking Detectors}
Watermark-based detection methods embed identifiable patterns or signals within generated text to confirm its origin. 
Early approaches, such as adversarial watermarking, directly embedded binary sequences into the content~\cite{watermark0:abdelnabi2021adversarial}, ensuring the watermark remained detectable. 
Recent methods focus on subtle watermarking by altering the model's sampling distribution to embed statistical traces without affecting fluency or semantics~\cite{Watermark1-forLLM:kirchenbauer2023watermark}. 
These techniques are useful in controlled environments where the text generator is known. 
However, watermarking is fragile under text manipulations like paraphrasing or compression~\cite{Retrieval:krishna2024paraphrasing}, which can invalidate the watermark. 
It is also ineffective for legacy outputs or open-source models that do not support watermark embedding~\cite{Watermark3-kirchenbauer2023reliability}, limiting its use in forensic scenarios. 
While watermarking has advantages in controlled settings, its limitations in open environments underscore the need for non-invasive detectors.

\subsubsection{Supervised Training-Based Methods}
In contrast to statistical feature-based methods, supervised training-based methods rely on machine learning models trained on labeled datasets to distinguish between human and machine-generated content. 
These methods have gained attention for their ability to learn complex patterns and improve detection accuracy in more challenging scenarios~\cite{yang2024survey}. 
For example, GPTZero~\cite{GPTZero} and OpenAI’s Classifier~\cite{OpenAI-CLS2019} use deep learning models, such as RoBERTa~\cite{OpenAI-CLS2019}, to classify text as human-written or machine-generated. 
These models are trained on large annotated datasets, distinguishing subtle differences between the two types of texts. 
However, they face issues with domain adaptation and cross-model generalization~\cite{hu2024radar}. 
These challenges are especially apparent when applied to out-of-domain texts or new models, as classifiers struggle with diverse text styles and maintaining robustness across domains~\cite{baseline3:RAIDAR:mao2024detecting}. 
Despite their strengths, supervised methods require regularly updated labeled datasets to remain effective in real-world scenarios.

\subsubsection{Statistical Feature-Based Methods}
Recent advancements in LLM-generated text detection have focused on enhancing robustness and generalization through statistical features, which quantify properties like log-likelihood ratios to distinguish human and machine-generated content. 
For example, DetectGPT~\cite{DetectGPT:mitchell2023detectgpt} detects machine-generated text by analyzing log-probability shifts when the original text is perturbed, identifying text with higher likelihoods under minor changes, a characteristic of LLMs.
This was optimized in Fast-DetectGPT~\cite{bao2024fast}, refining the perturbation process for better efficiency without compromising accuracy. 
These methods exploit the consistent likelihood patterns in machine-generated texts. 
Additionally, methods like BiScope~\cite{guo2024biscope} and Binoculars~\cite{baseline2:Binoculars:hans2024spotting} use cross-model perplexity contrasts, improving generalizability by comparing perplexity across models and using prediction discrepancies. 
However, challenges remain in detecting \textit{hybrid texts} (both human and machine-generated) and ensuring \textit{explainability}. 
Despite their promise, these methods struggle with the complexity of newer models.

\subsection{Advanced Detectors for Real-World Applications}
Recently, advanced detection methods have been developed to address real-world applicability challenges, particularly in detecting hybrid human-LLM content and ensuring real-time detection. 
For example, PALD~\cite{lei2025pald} estimates the proportion of machine-generated text and identifies specific sentences as machine-generated, making it useful for detecting partially generated content in academic papers or news articles.
Another key advancement is GLIMPSE~\cite{bao2025glimpse}, which bridges the gap between white-box and proprietary models by using partial observations to enable full probability distributions, enhancing detection across proprietary models.
Additionally, Song et al.~\cite{baseline1:R-Detect:song2025deep} employs a non-parametric kernel method to compare text distributions, improving detection accuracy and reducing false positives compared to traditional two-sample tests.
In parallel, the sequential hypothesis testing method~\cite{chen2025online} enables real-time detection by continuously analyzing streaming text, making it particularly well-suited for dynamic environments like social media and news outlets.
These methods represent significant progress in detection accuracy and real-world applicability, addressing challenges such as hybrid content detection, real-time analysis, and model generalization.

Content moderation and compliance applications face challenges in generalizing across domains and models, particularly in detecting hybrid content and ensuring explainability. 
To address this, we propose the \ourmethod framework, which redefines detection through the lens of \textit{stylistic divergence and style stability} under perturbations, offering a robust and explainable paradigm for identifying LLM-generated content.
While recent methods have improved accuracy, they often rely on task- or model-specific statistical cues and struggle with fine-grained, hybrid analysis of mixed-authorship texts.
\ourmethod bridges this gap by introducing a modular and scalable detection architecture that quantifies stylistic divergence at both global and local levels, enabling robust, explainable, and hybrid-aware detection across diverse domains.

\section{Methodology}
In this section, we define the problem of detecting machine-generated text and outline our approach for addressing it. 
Our method primarily focuses on leveraging stylistic features to distinguish between human-written and machine-generated content. 
We then introduce two types of stylistic features: discrete style features, which capture structural differences in the text, and continuous style features, which measure the semantic consistency across different texts. 
By combining these two feature types, our approach is designed to offer a robust and reliable solution for accurate text classification.

\begin{figure*}[!t]
    \centering
    \includegraphics[width=\linewidth]{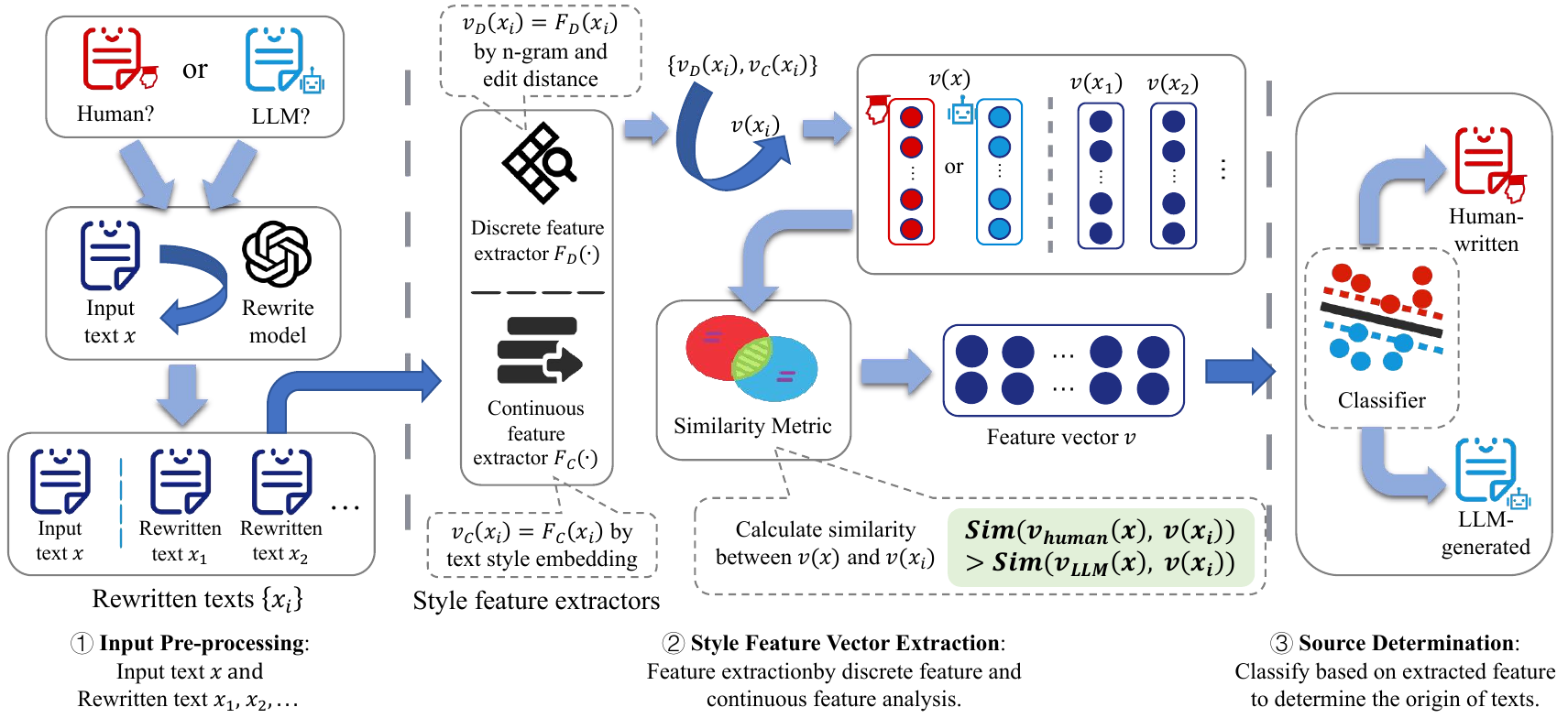}
    \caption{Overview of the proposed \ourmethod. 
    \ding{172} Given an input text $x$, a rewrite LLM model (GPT-3.5-turbo) generates a set of rewritten copies, denoted as $x^{\prime}_i$. 
    \ding{173} The original text $x$ and its corresponding rewritten copies are then channeled into our encoder. %feature extractor/embedder). 
    This extractor is designed to perform both discrete feature and continuous feature extraction. 
    Subsequently, the resulting feature vectors of these texts are compared using a similarity metric. 
    This comparative analysis yields a derived set of vectors $v$, which are specifically engineered for subsequent classification. 
    \ding{174} Processing this vector $v$ by a binary classifier to determine the source of the given text $x$.}
    \label{figure:scatter-plot}
\end{figure*}

\begin{table}[!t]
    \centering
    \normalsize
    % \small
    \caption{Main Notations}
    \label{table:notation}
    \renewcommand{\arraystretch}{1}
    \begin{tabular}{cp{0.73\linewidth}}
    \toprule
    \textbf{Notation} & \makecell[l]{\textbf{Explanation}} \\
    \midrule
    $x$ & Input text $x$ \\
    $y$ & Indicating the source of the text \\
    $S$ & Number of words in the text \\
    $L$ & Number of tokens in the text \\
    $t_i$ & The $i$-th token in the text \\
    $p$ & The list of prompts used for rewriting $p$ \\
    $\mathcal{M}$ & The LLM used for rewriting \\
    $\hat{x}$ & The rewritten text generated from $x$ \\
    $v$ & The feature vector of the text $x$ \\
    $v_D$ & The discrete style features \\
    $v_C$ & The continuous style stability features \\
    $v_N$ & The discrete features obtained from n-gram processing \\ % The discrete features obtained through n-grams
    $\operatorname{lev}_{x}(i,j)$ & The recursive function for edit distance computation; $i, j$ index first $i, j$ characters of $x, \hat{x}$ \\
    $v_{\text{edit}}$ & The discrete features obtained from edit distance analysis \\ % The discrete features based on Levenshtein edit distance
    $\xi$ & The text embedding model \\
    
    $\nu$ & Mapping of a single token to a vector representation \\
    $\mathbf{E}_i$ & the output of the $i$-th transformer layer\\
    $h$ & Number of layers in the encoder\\
    $H_i$ & the encoder of the $i$-th transformer layer \\
    $f$ & The classifier, taking $v$ as input and outputting $y_{\text{pred}}$ \\
    $y_{\text{pred}}$ & The predicted label of the text $x$ \\
    $N$ & Number of text samples \\
    $\mathcal{L}$ & Loss function \\
    $\Omega$ & Regularization term \\
    $\gamma, \lambda$ & Regularization parameters \\
    $\tau$ & Prediction threshold of the classifier \\
    \bottomrule
    \end{tabular}
\end{table}

\subsection{Problem Definition and Overview}
Machine-generated text detection is a binary classification problem where the goal is to determine the origin of a given text sequence \( x = \{ x_1, x_2, \dots, x_L \} \), where each \( x_i \) represents a token in the sequence. 
The text can either be generated by an LLM or written by a human. 
Let \( \mathcal{P} \) and \( \mathcal{Q}_\theta \) represent the distributions of human-written and machine-generated texts, respectively, over a metric space \( \mathcal{X} \).
Our goal is to classify a given text sequence \( x \) as either human-written or machine-generated by determining the most likely source of the text. 
Formally, we define the classification task as follows:
\begin{equation}
    \hat{y} = \underset{y \in \{0, 1\}}{\operatorname{argmax}} P (y \mid x, \mathcal{P}, \mathcal{Q}_\theta),
\end{equation}
where \( \hat{y} \in \{0, 1\} \) represents the predicted label.
The text \( x \) is a sequence of tokens \( x = [w_1, w_2, \dots, w_L] \), where each \( w_i \) is a token from the vocabulary. 
The objective is to detect if the given text has been generated by an LLM or is from human authorship based on stylistic patterns and robustness to perturbations.

\subsubsection*{Overview of the \ourmethod}
Our method leverages style divergence as the primary feature for detecting machine-generated text, based on the idea that LLM-generated text exhibits distinct stylistic patterns compared to human-written text. 
We analyze two key stylistic features for classification:
\textbf{\textit{(I) Discrete Style Features}}: We examine stylistic divergence between human and machine-generated texts using n-grams and edit distance.
\textbf{\textit{(II) Continuous Style Stability Features}}: We enhance detection by incorporating continuous style features that capture the stability of style across generations.
By combining discrete style features to identify stylistic inconsistencies and continuous style stability features to measure robustness against perturbations, our approach effectively captures both stylistic and semantic aspects of text generation.
\begin{tcolorbox}[colback=blue!3!white, colframe=blue!30!black, boxrule=0.6pt, arc=2mm, left=4pt, right=4pt, top=4pt, bottom=4pt]
    \textbf{\textit{Key Insight: Stylistic Divergence as a Universal Signal:}} \
    By combining discrete structural features and continuous semantic features, \ourmethod detects LLM-generated text by capturing stylistic differences between human and LLM-generated content while maintaining robust detection and strong explainability.
\end{tcolorbox}

\subsection{Discrete Style Features}
We begin by extracting discrete stylistic indicators from the input text \( x \), designed to capture token-level structural variation and invariance in style. 
These features include \( N \)-gram overlap and edit distance, measured between the original text and its perturbed version \( \hat{x} \), which is generated by a controlled rewriting process.
Given an input text \( x \) and a rewrite prompt \( p \), we generate a rewritten version \( \hat{x} \) using a language model \( \mathcal{M} \): $\hat{x} = \mathcal{M}(x, p)$.
This rewritten version maintains the semantic content of \( x \) while exhibiting stylistic variation. We compute discrete features by comparing \( x \) and \( \hat{x} \) across structural metrics.

\subsubsection{$N$-gram feature analysis}
$N$-gram is a contiguous sequence of \( N \) tokens from a given text. For a text \( x \) of length \( S \), we extract its \( N \)-grams as:
\begin{equation}
\begin{aligned}
    \mathcal{N}(n, x) = \{ (w_1, w_2, & \dots, w_n), (w_2, w_3, \dots, w_{n+1}), \\
    & \dots, (w_{S-n+1}, \dots, w_S)\}
\end{aligned}
\end{equation}
where \( S \) is the number of tokens in the text \( x \), and \( \mathcal{N}(n, x) \) is the set of all \( N \)-grams in \( x \). 
To capture surface-level stylistic consistency between the original and rewritten texts, we compute the normalized overlap between the \( N \)-gram sets of \( x \) and \( \hat{x} \) over a range of \( N \) values:
\begin{equation}
    v_N(x) = \frac{1}{n_2 - n_1} \sum_{n=n_1}^{n_2} \left| \mathcal{N}(n, x) \cap \mathcal{N}(n, \hat{x}) \right| .
\end{equation}
Here, \( n_1 \) and \( n_2 \) define the range of \( N \)-gram lengths, \( | \cdot | \) denotes set cardinality. 
Higher \( v_N(x) \) values suggest greater structural redundancy under rewriting.

\subsubsection{Levenshtein edit distance feature}
We also measure the normalized edit distance between \( x \) and its rewritten counterpart \( \hat{x} \) using Levenshtein distance computed via dynamic programming~\cite{baseline3:RAIDAR:mao2024detecting}:
\begin{equation}
    \mathcal{L}(x, \hat{x}) = N_{\text{min} | x \rightarrow{} \hat{x}},
\end{equation}
where $N_{\text{min} | x \xrightarrow{} \hat{x}}$ means the minimum number of edits to convert $x$ to $\hat{x}$. 
Let $\mathcal{L}(x, \hat{x}) = \mathcal{L}_{x}(|x|, |\hat{x}|)$, and it can be calculated through the following iterative approach:
\begin{equation}
\small
\begin{aligned}    
    \mathcal{L}_{x}(i, j) =
    \begin{cases}
        \max(i, j),   & \text{if $\min(i, j) = 0$,} \\ 
        \min   \begin{cases} 
                \mathcal{L}_{x}(i - 1, j) + 1, \\
                \mathcal{L}_{x}(i, j - 1) + 1, \\
                \mathcal{L}_{x}(i-1, j-1) + 1_{i,j},
    \end{cases} & \text{otherwise.}
    \end{cases}
\end{aligned}
\end{equation}
where $1_{i,j}$ is the indicator, which is equal to $0$ when $x_i=\hat{x}_j$, otherwise it is 1.
$\mathcal{L}_{x}(i, j)$ represents the edit distance between the first $i$ characters of $x$ and the first $j$ characters of the rewritten version $\hat{x}$.
Then the edit distance is defined as:
\begin{equation}
    v_{\text{edit}}(x) = 1 - \frac{\mathcal{L}(x, \hat{x})}{\max(S_x, S_{\hat{x}})},
\end{equation}
where \( S_x \) and \( S_{\hat{x}} \) denote the number of words in \( x \) and \( \hat{x} \), respectively. 
This feature captures character-level transformations induced by style-preserving rewriting.
Therefore, we can concatenate the \( N \)-gram and edit-distance-based features to get the discrete feature vector:
\begin{equation}
    v_D(x) = \{v_N(x), v_{\text{edit}}(x)\}.
\end{equation}
This vector encapsulates token-level stylistic stability and forms a critical input to our final classification model.

\subsection{Continuous Style Stability Features}
In addition to discrete style features, we compute continuous style stability features using text embeddings. We expect to utilize text embeddings to capture the semantic characteristics of the text and its stability across different perturbations. 

% \subsubsection{BERT-based embedding model}
The used embedded model is trained based on the BERT model.
% So we consider the processing procedure of the BERT model. 
The original input text $x$ is first divided by the WordPiece tokenizer $W$ into tokens $W(x)= \{t_1, t_2, \dots, t_L \}$, which are from the vocabulary $A$.
BERT's vocabulary $A$ contains all the tokens that the model can handle. 
Once the text is decomposed into its tokens, we can use the built-in vocabulary of the BERT model to find the embedding vectors for each smallest unit.
For each token $t_j \in \{t_1, t_2, \dots, t_L \}$, the model assigns a unique index to it and searches for the corresponding embedding vector $\nu^{(1)}(t_j) \in  \mathbb{R}^d$ from the embedding matrix based on the index.

For each token \( t_j \), we add a positional encoding \( PE \) to represent its position in the sentence (where $j$ is the token’s index):
\begin{equation}
    \hat{\nu}^{(2)}(t_j) = \nu^{(1)}(t_j) + PE(j)
\end{equation}

After adding the positional encoding, we obtain the input embedding sequence $\mathbf{E}_{0} = [\hat{\nu}^{(2)}(t_{1}); \hat{\nu}^{(2)}(t_{2}); \dots; \hat{\nu}^{(2)}(t_{L})]$ (a matrix of $L \times d$ dimensions), which is then fed into the Transformer encoder \( {H} \) of BERT:
\begin{equation}
    \mathbf{E}_{i} = {H}_{i}(\mathbf{E}_{i-1}), \quad \text{for } i=1, \dots, h
\end{equation}
where $h$ is the total number of layers in the encoder, ${H}_ i$ is the encoder of the $i$-th layer, \( \mathbf{E}_{i} \) is the output of the \( i \)-th layer, and \( \mathbf{E}_{0} \) represents the input embedding sequence.

Let \( \xi \) denote the entire model (including the tokenizer and transformer encoder). 
Then, we can represent the embedding process as:
\begin{equation}
    \xi(x) = \mathbf{E}_{h} \in \mathbb{R}^{L \times d}.
\end{equation}
% \subsubsection{cosine similarity}
To obtain a single fixed-size vector representation for the entire text, we apply a pooling operation on $\xi (x)$, resulting in a text embedding vector $ \hat \xi(x) \in \mathbb{R} ^ d$. 
 Then the cosine similarity between the text embedding
vector of $x$ and the text embedding vector of the rewritten $\hat x$ can be calculated as:
\begin{equation}
    v_C(x) = \frac{\hat \xi(x) \cdot \hat\xi(\hat{x}) }{\left\| \hat\xi(x) \right\| \left\|\hat\xi(\hat{x}) \right\|}
\end{equation}

\subsection{Detection via Stylistic Feature Divergence}
To classify the input text, we combine the discrete style features and continuous style stability features into a single detection feature \( v(x) \):
\begin{equation}
    v(x) = \alpha \cdot v_D(x) \oplus \beta \cdot v_C(x),
\end{equation}
where \( v_D(x) \) is the discrete style feature vector, \( v_C(x) \) is the continuous style stability feature vector, and \( \alpha, \beta\) are hyperparameters.

Therefore, we can obtain the feature representations $V = \left\{v\left(x_1\right), v\left(x_2\right), \ldots, v\left(x_N\right)\right\}$ of the set of samples $X = \{x_1, x_2, \dots, x_N\}$ (associated labels $Y = \{y_1, y_2, \dots, y_N\}$).
Then, we can get the prediction $y_{\text{pred}} = f(x)$ by a text evaluation classifier $f$.
We implement $f$ as the XGBoost trained by the Gradient Boosting in this work.
Consider a training sample $\left\{\left(x_i, y_i\right)\right\}$, where $\hat{y}_i$ denotes the prediction.
Let $K$ denote the total number of trees in the ensemble model. 
Each tree $f_k\left(x_i\right)$ represents the $k$-th tree’s prediction output on the sample $x_i$. 
The model comprises \( K \) trees, each represented as a base learner \( f_k(x_i) \) contributing to the prediction process, where \( k \) is the index. 
\begin{algorithm}[!t]
    \caption{\ourmethod: LLM-Generated Text Detection via Stylistic Features}
    \label{alg:styledecipher}
    \begin{algorithmic}[1]
        \STATE \textbf{Input}: Input text $x$, rewriting model $\mathcal{M}$, rewriting prompt $p$, style encoder $\xi$, classifier $f$, thresholds $\alpha$, $\beta$, and decision threshold $\tau$
        \STATE \textbf{Output}: Prediction $\hat{y} \in \{0, 1\}$ of whether $x$ is LLM-generated
        % \STATE \textbf{// Step 1: Style-Preserving Rewriting}
        \STATE Use rewriting model $\mathcal{M}$ with prompt $p$ to generate rewritten version $ \hat{x} = \mathcal{M}(x, p) $
        % \STATE \textbf{// Step 2: Discrete Style Feature Extraction}
        \STATE Compute $N$-gram sets $\mathcal{N}(n, x)$ and $\mathcal{N}(n, \hat{x})$ for $n \in [n_1, n_2]$ to obtain the $N$-gram feature overlap:
        $$ v_N(x) = \frac{1}{n_2 - n_1} \sum_{n=n_1}^{n_2} \left| \mathcal{N}(n, x) \cap \mathcal{N}(n, \hat{x}) \right| $$
        \STATE Compute normalized edit distance via dynamic programming:
        $$ v_{\text{edit}}(x) = 1 - \frac{\mathcal{L}(x, \hat{x})}{\max(S_x, S_{\hat{x}})} $$
        \STATE Concatenate discrete features: $v_D(x) = \{v_N(x), v_{\text{edit}}(x)\}$
        % \STATE \textbf{// Step 3: Continuous Style Stability Feature Extraction}
        \STATE Encode original and rewritten texts with style encoder: $\xi(x), \; \xi(\hat{x}) \in \mathbb{R}^{L \times d}$
        \STATE Apply a pooling operation on $\xi(x)$ and $\xi(\hat{x})$ to obtain text embedding vectors $\hat{\xi}(x), \hat{\xi}(\hat{x}) \in \mathbb{R}^d$.
        \STATE Compute continuous style stability feature:
        $$ v_C(x) = \frac{\hat \xi(x) \cdot \hat\xi(\hat{x}) }{\left\| \hat\xi(x) \right\| \left\|\hat\xi(\hat{x}) \right\|} $$
        % \STATE \textbf{// Step 4: Feature Fusion and Classification}
        \STATE Combine discrete and continuous features:
        $$ v(x) = \alpha \cdot v_D(x) \oplus \beta \cdot v_C(x) $$
        % \textbf{// Step 5: Thresholding Decision}
        \STATE  Predict detection probability and logit using classifier $f$ (e.g., XGBoost) and the threshold $\tau$:
        $$ \hat{y}_{\text{logit}} = f(v(x)), \quad p = \sigma(\hat{y}_{\text{logit}}) = \frac{1}{1 + e^{-\hat{y}_{\text{logit}}}} $$
        \STATE \textbf{Return} $\hat{y} = \mathbb{I}[p > \tau]$
    \end{algorithmic}
\end{algorithm}

During training, the model is updated in \( K \) rounds. 
In each iteration, a new decision tree is trained to correct the residual errors from the previous round. 
Specifically, at iteration \( t \), for each sample \( i \), the prediction at iteration \( t \) is computed as:
\begin{equation}
    \hat{y}_i^{(t)} = \hat{y}_i^{(t-1)} + \eta \cdot f_t(x_i)
\end{equation}
where \( f_{t}(x_i) \) represents the current model's prediction for sample \( i \), and \( \eta \) is the learning rate.

The object of iteration \( t \) is to minimize the regularized loss, so we consider the loss function as:
\begin{equation}
    \mathcal{L}^{(t)} = - \sum_{i=1}^N [y_i \log(p_i) + (1 - y_i) \log(1 - p_i)] + \sum_{k=1}^t \Omega(f_k).
\end{equation}
% The loss function \( L(y_i, \hat{y}_i^{(t)}) \) is the standard binary cross-entropy:
where the predicted probability \( p_i \) is obtained through:
\begin{equation}
    p_i = \sigma(\hat{y}_i) = \frac{1}{1 + e^{-\hat{y}_i}}
\end{equation}
Here, the regularization term \( \Omega(f_k) \) is a regularization term on the complexity of the \( k \)-th tree, used to control its size and avoid overfitting.
\begin{equation}
    \Omega(f) = \gamma T + \frac{1}{2} \lambda \sum_{j=1}^T w_j^2
\end{equation}
where \( T \) is the number of leaf nodes in the tree, \( w_j \) is the weight of the \( j \)-th leaf, and \( \gamma \), \( \lambda \) are regularization parameters.

To optimize the objective function, we compute the first and second-order derivatives of the loss for \( \hat{y}_i^{(t-1)} \), denoted as \( g_i^{(t)} \) and \( h_i^{(t)} \) respectively:  % (Hessian matrixs)
\begin{equation}
    g_i^{(t)} = \frac{\partial L(y_i, F^{(t-1)}(x_i))}{\partial F^{(t-1)}(x_i)},
\end{equation}
\begin{equation}
    h_i^{(t)} = \frac{\partial^2 L(y_i, F^{(t-1)}(x_i))}{\partial (F^{(t-1)}(x_i))^2}.
\end{equation}
Therefore, the loss function is approximated as:
\begin{equation}
    \mathcal{L}^{(t)} \approx \sum_{i=1}^N \left[ g_i^{(t)} f_t(x_i) + \frac{1}{2} h_i^{(t)} (f_t(x_i))^2 \right] + \Omega(f_k)
\end{equation}
By optimizing this objective, we obtain the best-fitting tree \( f_t \), and update the model prediction as:
\begin{equation}
    \hat{y}_i^{(t)} = \hat{y}_i^{(t-1)} + \eta \cdot f_t(x_i).
\end{equation}
After \( K \) iterations, the final ensemble consists of \( K \) additive trees \( \{f_1, f_2, \ldots, f_K\} \).
For a given test sample \( x_{\text{test}} \), the model outputs a logit prediction:
\begin{equation}
    \hat{y}_{\text{test}} = \hat{y}^{(0)} + \eta \sum_{k=1}^K f_k(x_{\text{test}})
\end{equation}
Therefore, the detection probability $\sigma(\hat{y}_{\text{test}})$ can be obtained and the detection decision is made based on the threshold \( \tau \).

We give the outline of the complete process of our proposed \ourmethod detection framework in Algorithm~\autoref{alg:styledecipher}, which integrates rewriting, stylistic feature extraction, and classification into a unified pipeline. 
The algorithm takes as input a raw text sample $x$, a rewrite model $\mathcal{M}$, a set of rewriting prompts $p$, a style embedding encoder $\xi$, and a binary classifier $f$ trained on feature vectors. 
The process begins by generating a rewritten version $\hat{x} = \mathcal{M}(x, p)$, which preserves semantic content while introducing stylistic variation. 
We then extract two types of stylistic features: (I) discrete style features $v_D(x)$, including $N$-gram overlap and normalized edit distance between $x$ and $\hat{x}$, and (II) continuous style stability features $v_C(x)$, calculated via cosine similarity between the embedding vectors of $x$ and $\hat{x}$. 
These features are fused into a unified representation $v(x) = \alpha \cdot v_D(x) \oplus \beta \cdot v_C(x)$, where $\alpha$ and $\beta$ are hyperparameters balancing the contribution of each feature type. 
The final prediction $\hat{y}$ is made by passing $v(x)$ through the classifier $f$ and comparing the output probability $\sigma(\hat{y})$ to a decision threshold $\tau$. 
% If $\sigma(\hat{y}) > \tau$, the text is classified as LLM-generated; otherwise, it is labeled as human-written.

\section{Experiments}
In this section, we first introduce the experimental setup. 
Then, the robustness performance is evaluated, covering in-domain accuracy, cross-domain generalization and performance under adversarial perturbations and data mixing. 
Finally, we assess the model's explainability and flexibility. 

\subsection{Evaluation Setup}
\subsubsection{Datasets and LLMs}
We evaluate our method on five datasets from diverse domains, each containing equal numbers of human-written and LLM-generated texts. 
The LLM-generated texts are produced by prompting advanced LLMs with the same topics or prompts as the human-written texts. 
The evaluation spans various genres, including formal writing, student essays, programming code, customer reviews, and scientific abstracts.
A summary of datasets is provided below:
\begin{itemize}
    \item \textit{News:} The human-written texts are sampled from the \text{Reuter\_50\_50} dataset~\cite{houvardas2006n}, originally compiled in 2006. 
    We use the version released in~\cite{baseline4:Ghostbuster:verma2024ghostbuster}, which generated corresponding LLM-written texts using ChatGPT (davinci). 
    
    \item \textit{Student Essays:} Human-written essays are collected from IvyPanda~\cite{ivypanda}, a repository of student-written essays. 
    Following~\cite{baseline4:Ghostbuster:verma2024ghostbuster}, the LLM-generated counterparts were created using ChatGPT. % (davinci variant). 
    
    \item \textit{HumanEval Code:} We use the HumanEval dataset~\cite{baseline3:RAIDAR:mao2024detecting} as the source of human-written code snippets. 
    The LLM-generated code is obtained by prompting GPT-3.5-turbo, following the procedure introduced in~\cite{baseline3:RAIDAR:mao2024detecting}.
    
    \item \textit{Yelp Reviews:} This dataset is adopted from~\cite{baseline3:RAIDAR:mao2024detecting}. 
    Human-written reviews are collected from Yelp~\cite{zhang2015character}.
    Their LLM-generated counterparts are synthesized using \textit{GPT-3.5-turbo}, following \cite{baseline3:RAIDAR:mao2024detecting}. 
    % Although the precise crawl time is not specified, its validity is supported by direct reuse by other recent studies, including~\cite{guo2024biscope}.
    
    \item \textit{Paper Abstracts:} Human-written texts are drawn from ACL 2023 papers, where 500 abstracts are sampled. 
    The LLM-generated texts are produced using the same methodology as~\cite{baseline3:RAIDAR:mao2024detecting}, with GPT-3.5-turbo conditioned on paper titles.

    \item \textit{RAID Dataset:} RAID~\cite{dugan2024raid} is the largest and most comprehensive dataset for evaluating AI-generated text detectors, which contains over 10 million documents spanning 11 LLMs, 11 genres, 4 decoding strategies, and 12 adversarial attacks. 
\end{itemize}

\subsubsection{Baselines}
We compare \ourmethod with state-of-the-art text detection methods:
\begin{itemize}
    \item \textit{GPTZero}~\cite{GPTZero} is a commercial classifier relying on handcrafted features and syntactic heuristics for detecting LLM-generated texts. 
    It serves as a strong practical baseline for detecting \textit{ChatGPT}, \textit{LLaMa}, and other LLMs. 
    
    \item \textit{DetectGPT}~\cite{DetectGPT:mitchell2023detectgpt} identifies LLM-generated text by examining changes in curvature of log-probability under small input perturbations. 
    It relies on model scoring functions and assumes internal access to a reference LLM. 

    \item \textit{Ghostbuster}~\cite{baseline4:Ghostbuster:verma2024ghostbuster} is a black-box detector that ensembles features from multiple weaker LMs and achieves strong cross-domain generalization without accessing target model token probabilities. 

    \item \textit{RAIDAR}~\cite{baseline3:RAIDAR:mao2024detecting} detects AI-generated text by prompting LLMs to rewrite inputs and computing edit distances, leveraging the observation that LLMs alter human-written text more than their own outputs. 

    \item \textit{Fast-DetectGPT}~\cite{bao2024fast} is an efficient zero-shot detector that estimates probability curvature through conditional sampling instead of perturbation. 
    It significantly improves detection speed while retaining strong performance under both white-box and black-box conditions. 

    \item \textit{Binoculars}~\cite{baseline2:Binoculars:hans2024spotting} is a zero-shot detector that contrasts log-probability outputs from two similar pre-trained LLMs, requiring no fine-tuning or training data and performing well across diverse text types. 

    \item \textit{R-Detect}~\cite{baseline1:R-Detect:song2025deep} applies a non-parametric kernel relative test to determine whether a test text is statistically closer to human or machine distributions, effectively reducing false positives in hybrid or outlier cases. 
\end{itemize}
% ---------------
\begin{figure*}[!th]
    \centering
    \includegraphics[width=\linewidth]{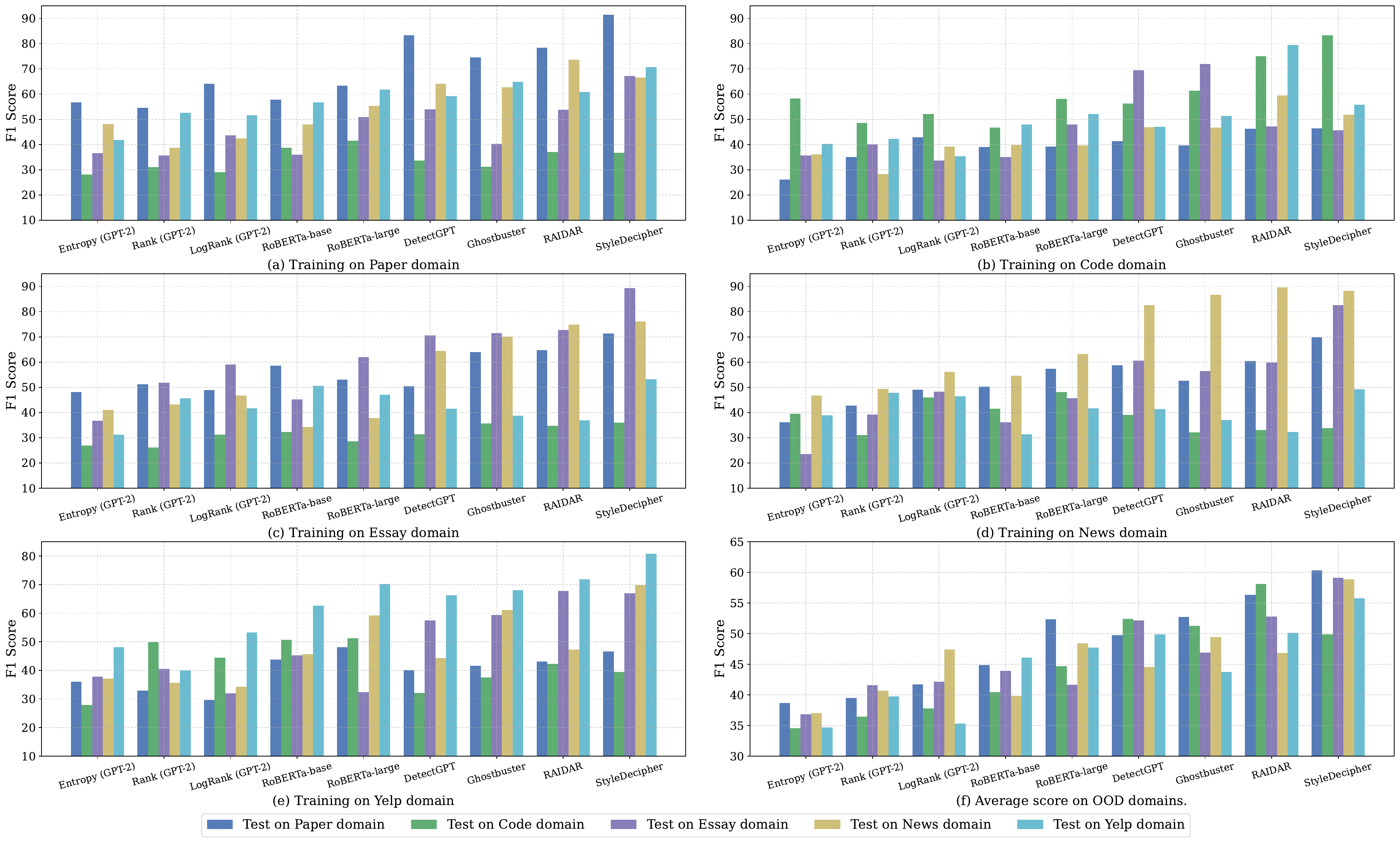}
    \caption{Cross-domain generalization F1 score on ID-OOD splits. 
    Each subfigure represents the results on a source training domain, where each bar shows the result on the target evaluation domain. 
    The last subfigure denotes the average F1 score on out-of-domain targets.
    }
    \label{figure:OOD}
\end{figure*}
% --------------------

\subsubsection{Implementation}
Baseline methods like Fast-DetectGPT, Binoculars, and R-Detect are similar to the previous DetectGPT, all adopting a threshold-setting approach. 
In our main comparative experiments, we use AUROC as the metric to improve their performance, ensuring fairness in the experiments. 
Regarding threshold setting, we follow the official implementations of Binoculars and R-Detect.
For Ghostbuster, the original work used log probabilities from GPT Ada and Davinci models, but these two models are now nearly deprecated. In this paper, we replace them with \textit{GPT-3.5-Turbo-1106} and \textit{GPT-3.5-turbo-16k}.
For DetectGPT, we follow the original paper and use T5-3B as the perturb model. 
To ensure fairness, the F1 score results are computed by traversing the thresholds to select the optimal result.

\subsection{In-Domain and Cross-Domain Detection Performance}
\subsubsection{In-Domain Detection Accuracy}
\begin{table}[!t]
  \centering
  % \small
  \footnotesize
  \setlength{\tabcolsep}{2.0pt} 
  \caption{Main detection AUROC scores of the texts generated by the \GPTThreeFiveTurbo across four domains.}
  \label{table:Main-deteection}
  \resizebox{\linewidth}{!}{
  \begin{tabular}{lcccccc}
    \toprule
    \textbf{Method} & News & HumanEval & Essay & Yelp Review & \textbf{Avg} \\  % & Writing
    \midrule
    Entropy (GPT-2) & 0.4246 & 0.4306 & 0.4808 & 0.4697 & 0.4514 \\
    Rank (GPT-2) & 0.6560 & 0.5348 & 0.6849 & 0.6819 & 0.6394 \\
    LogRank (GPT-2) & 0.7438 & 0.5350 & 0.6758 & 0.5294 & 0.6210 \\
    RoBERTa-base & 0.7024 & 0.4217 & 0.6317 & 0.4723 & 0.5570 \\
    RoBERTa-large & 0.7301 & 0.4692 & 0.3325 & 0.4061 & 0.4845 \\
    DetectGPT~\cite{DetectGPT:mitchell2023detectgpt} & 0.8213 & 0.5267& 0.6410 & 0.6342 &  0.6558\\
    Ghostbuster~\cite{baseline4:Ghostbuster:verma2024ghostbuster} & 0.6401 & 0.5378 & 0.5798 & 0.6691 &  0.6067\\
    Fast-DetectGPT~\cite{bao2024fast} &  0.9486 & 0.6679 & 0.9206 & 0.6230&  0.7923\\
    RAIDAR~\cite{baseline3:RAIDAR:mao2024detecting} & 0.8956 & 0.8173 & 0.9091 & 0.8616 & 0.8709 \\
    % Bino(2024) & \textbf{0.9944}& 0.7678 & \textbf{0.9827} & 0.6821&  0.8568\\
    R-Detect~\cite{baseline1:R-Detect:song2025deep} & 
    \textbf{0.9817} & 0.6490 &  0.7629 & 0.7121  &  0.7764 \\
    \midrule
    \ourmethod & 0.9356 & \textbf{0.8108} &  \textbf{0.9455} & \textbf{0.8718} & \textbf{0.8927}\\
    \bottomrule
    \end{tabular}
    }
\end{table}
To investigate whether stylistic divergence and stability can serve as reliable signals for detecting LLM-generated texts, we evaluate the detection performance of various methods across multiple representative domains. 
As shown in Table~\ref{table:Main-deteection}, \ourmethod achieves the highest or near-highest AUROC scores in all tested domains, demonstrating its strong in-domain detection capabilities. 
Particularly in structurally diverse domains such as code and student essays, our approach outperforms state-of-the-art baselines like DetectGPT and Ghostbuster, which often struggle to capture nuanced stylistic deviations inherent to human writing. 
Traditional statistical detectors (e.g., entropy and rank-based methods), although lightweight and model-agnostic, consistently underperform due to their inability to effectively model high-level stylistic coherence. 
Similarly, supervised detectors such as RoBERTa exhibit inconsistent performance, likely due to a domain mismatch between training and test data.

% -------------------
\begin{figure*}[!t]
    \centering
    \includegraphics[width=\linewidth]{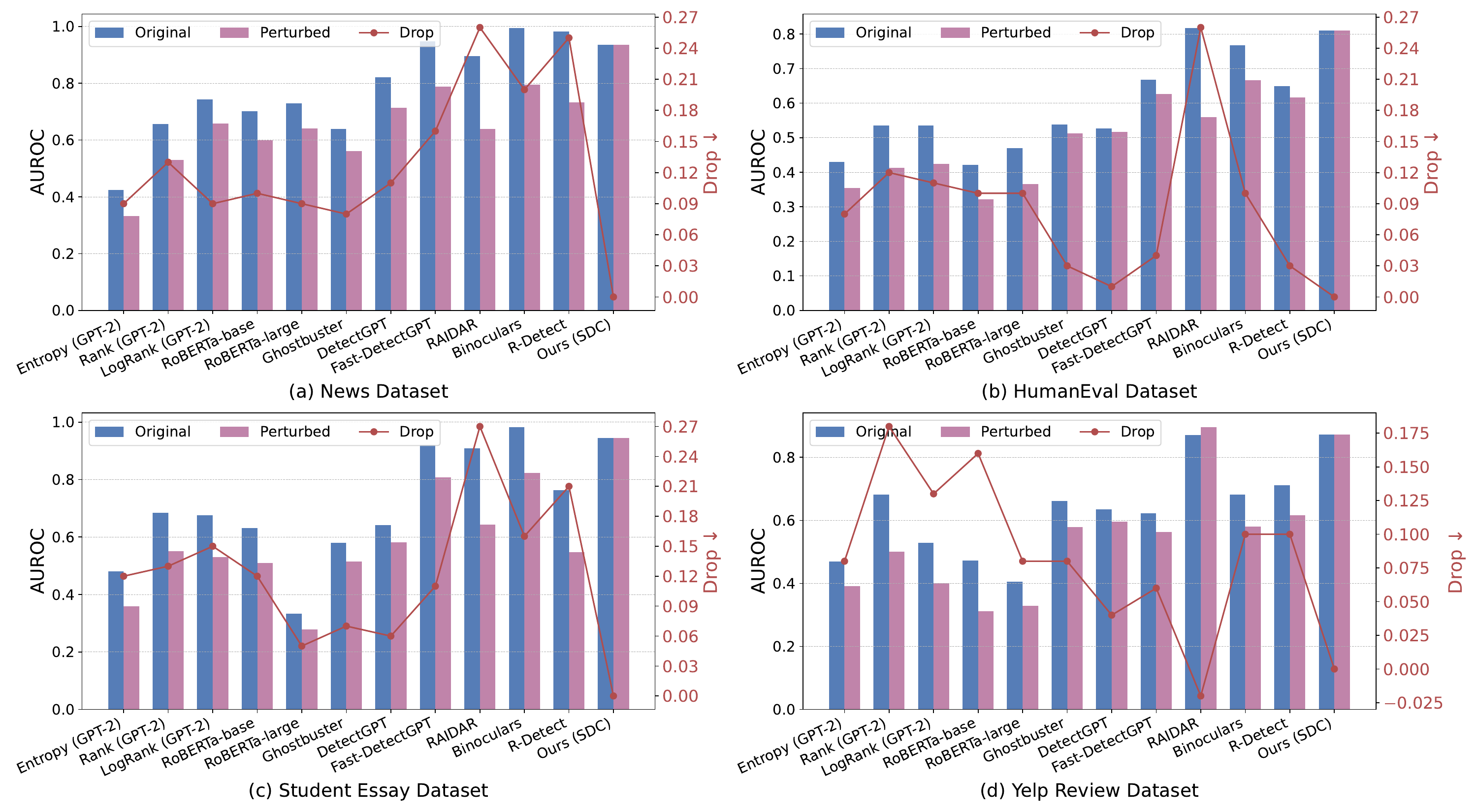}
    \caption{Evaluation of detection performance degradation under adversarial perturbations: Comparative analysis of original AUROC, AUROC after perturbation, and relative drop rate across News, HumanEval, Code, and Yelp Review datasets.}
    \label{figure:perturbation}
\end{figure*}
% -------------------

Unlike methods that rely on token-level scoring or access to LLM internals, \ourmethod utilizes the divergence between original and rewritten texts as a core signal, which is captured via both discrete structural features (n-gram overlap, edit distance) and continuous semantic embeddings (BERT-based similarity). 
This hybrid representation allows for effective separation between human-written and machine-generated content, even under stylistic noise or domain shifts. 
Furthermore, our modular design enhances robustness and adaptability without overfitting to specific genres. 
The method consistently maintains high detection accuracy while offering explainable predictions, addressing the dual objectives of robustness and explainability for RQ1 and RQ2. 
This consistency across domains, where other detectors show marked performance degradation, demonstrates the generalizability of stylistic perturbation analysis as a model-agnostic detection strategy. 

\subsubsection{Cross-Domain Generalization Ability}
As demonstrated in~\autoref{figure:OOD}, the cross-domain generalization performance of our model was evaluated through leave-one-domain-out experiments. 
Each subfigure represents the results of training on one source domain and evaluating it on all five target domains. 
The results clearly indicate that \ourmethod consistently outperforms the baseline methods, including RAIDAR, across various domain transfer settings, which highlights the robustness and adaptability of our model in handling diverse text types, making it highly effective in real-world applications.

(I) \textbf{Training on Formal Domains}: When trained on more formal domains, such as Paper and Essay domains (\autoref{figure:OOD} (a) and (c)), our model maintains a high F1 score when evaluated on more diverse and stylistically distinct domains like News and Yelp reviews.
For instance, in \autoref{figure:OOD} (a) (training on Paper), \ourmethod shows robust performance on domains such as News and Code, outperforming other methods like RAIDAR.
(II) \textbf{Training on Casual Domains}: In \autoref{figure:OOD} (e) and (f), where the model was trained on Yelp reviews, \ourmethod demonstrates exceptional adaptability, maintaining high detection performance even when transferred to more structured domains like Code and Paper. 
This suggests that stylistic divergence enables our model to generalize well, even when trained on less formal or more conversational domains.
(III) \textbf{Cross-Domain Performance}: \autoref{figure:OOD} (f) highlights the average performance on out-of-domain (OOD) targets, with \ourmethod achieving higher average F1 scores than all other baseline methods. 
This reinforces the conclusion that our approach is not limited by specific text types or writing styles, but rather can effectively detect machine-generated text across a wide variety of genres and domains.
(IV) \textbf{Outperformance on Complex Cross-Domain Transfers}: \ourmethod excels at challenging domain transfers, such as detecting text in natural language content format when the model is trained on code, as shown in \autoref{figure:OOD} (b) and (d).
This result further supports the idea that the stylistic divergence framework captures generation-invariant patterns, which are applicable across different domains, thus ensuring high transferability and adaptability to diverse text types.

Overall, the results presented in~\autoref{figure:OOD} confirm that \ourmethod is highly effective in real-world scenarios, where text comes from diverse sources with varying stylistic features. Unlike previous methods, which may struggle with domain-specific heuristics, our model's reliance on stylistic divergence provides a robust solution to detecting LLM-generated text, regardless of the text's genre or domain.
\begin{tcolorbox}[colback=blue!3!white, colframe=blue!30!black, boxrule=0.6pt, arc=2mm, left=4pt, right=4pt, top=4pt, bottom=4pt]
\textbf{\textit{Takeaway – Domain-Robust Detection:}} \\
(I) Our method maintains stable in-domain performance, outperforming baselines, especially in complex domains like papers, essays, and code. \\
(II) It generalizes well under domain shifts, with strong cross-domain results, showing that stylistic divergence captures transferable signals, even when trained on informal domains like Yelp reviews.
\end{tcolorbox}

% -------------------
\begin{figure*}[!t]
    \centering
    \includegraphics[width=\linewidth]{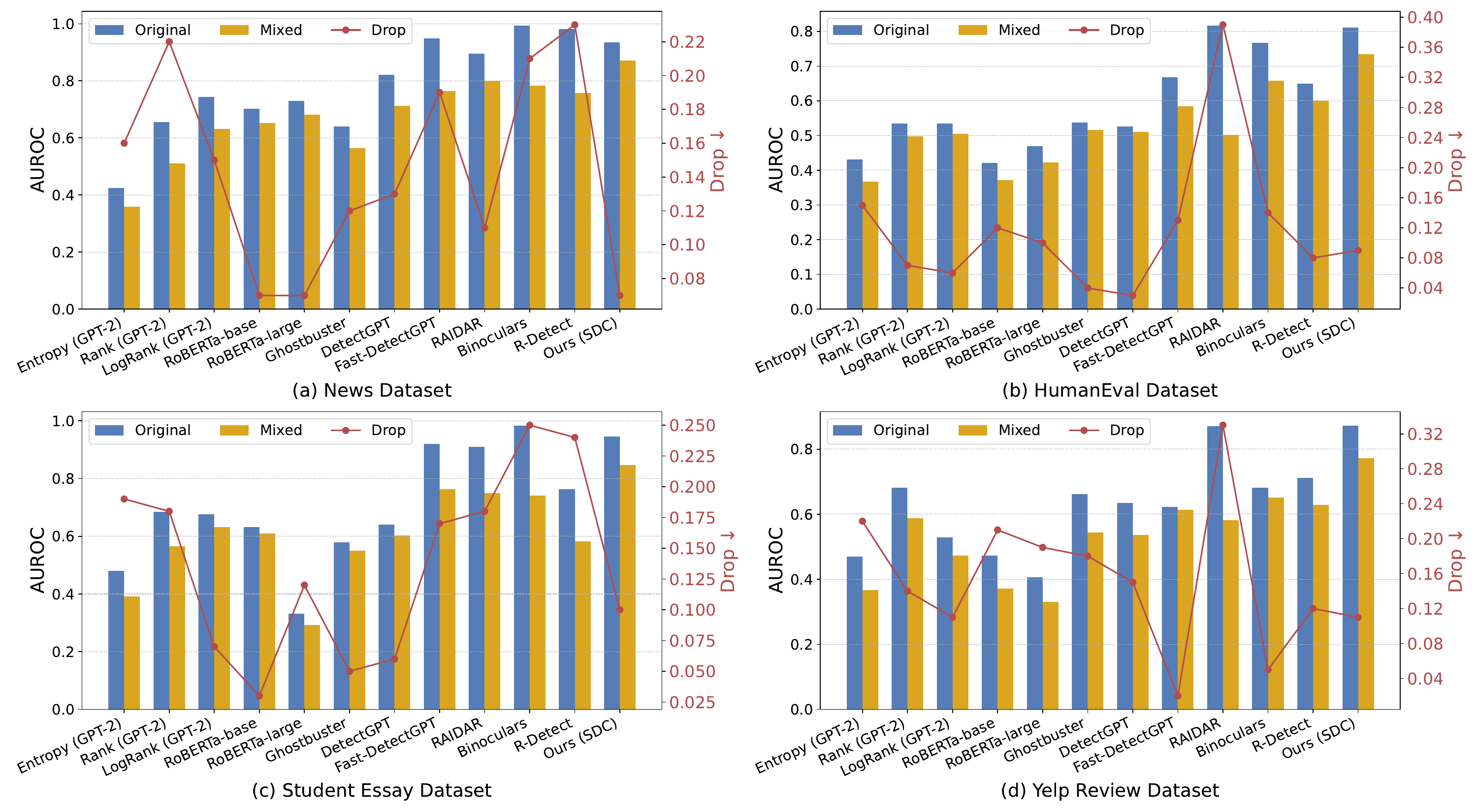}
    \caption{Evaluation of detection performance degradation under adversarial mixed text: Comparative analysis of original AUROC, AUROC after mixing, and relative performance drop across News, HumanEval, Code, and Yelp Review datasets.}
    \label{figure:mixed}
\end{figure*}
% -------------------------

\subsection{Adversarial Robustness to Perturbations and Data Mixing}
\subsubsection{Resilience to Adversarial Perturbations}
The robustness of our model against adversarial perturbations is evaluated as shown in ~\autoref{figure:perturbation}), where the performance degradation of different models under adversarial attacks is visualized, specifically implemented by \textit{TextAttack}. 
The comparison of perturbed AUROC, original AUROC, and the relative performance drop across four datasets, including News, HumanEval, Code, and Yelp Review, provides an understanding of how well each model withstands perturbations relative to its original performance. 
The bars represent the AUROC scores for the original and perturbed inputs, while the red line illustrates the relative drop in performance, highlighting the models' resilience.

From the results, we observe that \ourmethod consistently outperforms the baselines under adversarial perturbations, particularly when we examine the relative performance drop across datasets. 
For instance, while models like DetectGPT and Ghostbuster suffer from significant drops in performance, especially in domains like Code and Yelp Review, \ourmethod maintains relatively stable performance even under these challenging conditions. 
In Yelp Review, where the perturbations cause a noticeable AUROC decrease of over 10\% for other models, \ourmethod stands out with a slight improvement in performance, demonstrating its strength in handling real-world text manipulations. 
Similarly, while other methods experience substantial declines in AUROC in domains such as Student Essays and News, \ourmethod exhibits a much smaller reduction, reinforcing its robustness.
These results directly address the RQ1 by demonstrating that \ourmethod is not only effective at detecting machine-generated text in clean datasets but also resilient under adversarial conditions. 
The relative drop line, which tracks the performance degradation, underscores the effectiveness of \ourmethod in maintaining high detection performance in adversarial settings. 
This suggests that the model captures more fundamental, generation-invariant stylistic signals, allowing it to withstand manipulations designed to exploit simpler detection methods. 
The model's robustness across a variety of datasets, including those most susceptible to adversarial attacks, highlights its practical applicability for deployment in real-world detection systems where adversarial attacks are increasingly common.

% -----------------------------
\begin{figure*}[!t]
    \centering
    \footnotesize
    \setlength{\tabcolsep}{2.0pt}
    \includegraphics[width=\linewidth]{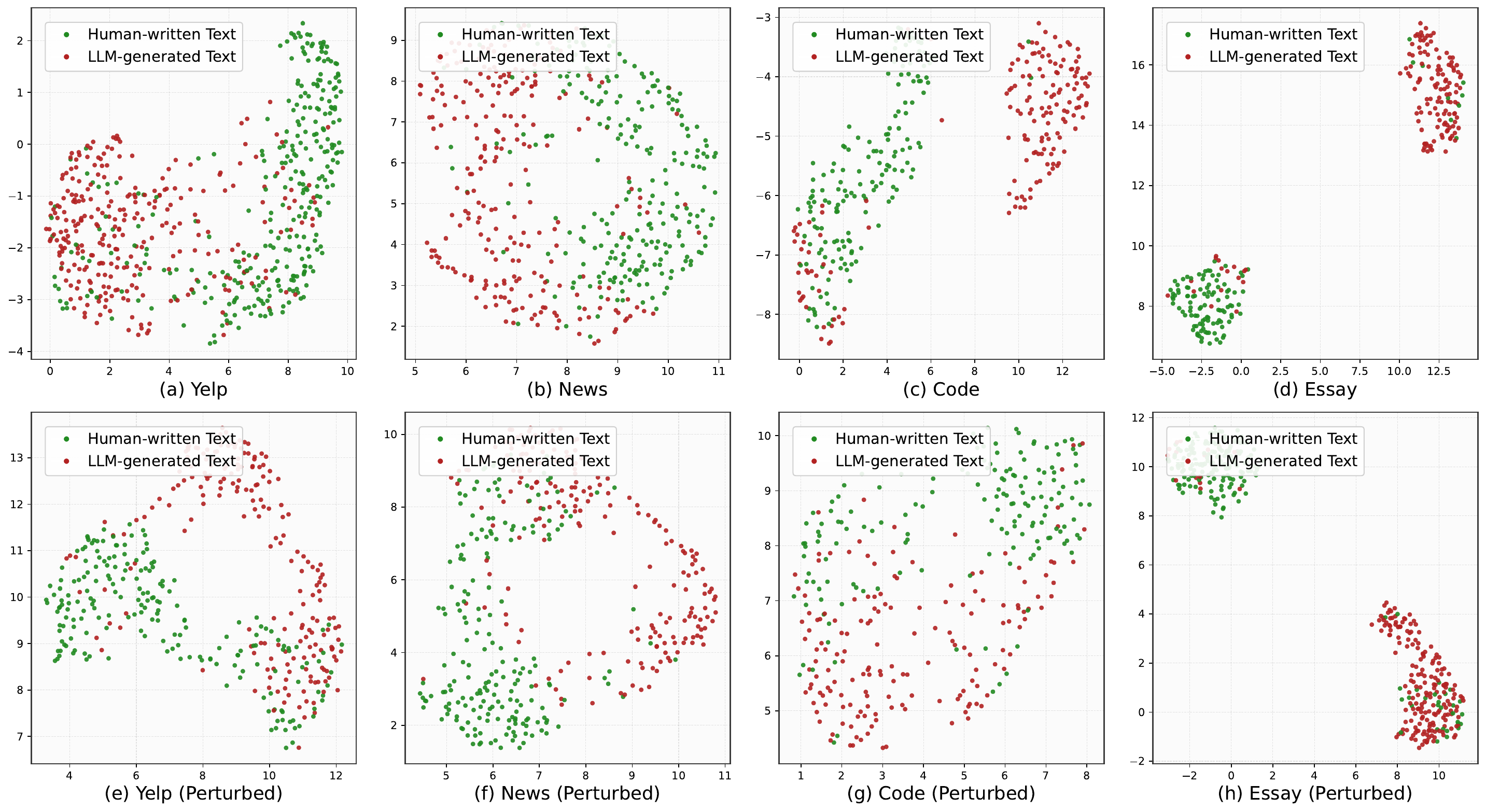}
    \caption{Explainable visualization of in-domain and perturbed text detection: UMAP projection of human-written and LLM-generated texts across multiple domains (Yelp, News, Code, Essay).}
    \label{figure:explainability-UMAP}
\end{figure*}
% -------------------------------

\subsubsection{Performance on Adversarial Data Mixing}
The robustness performance under adversarial mixing is assessed in \autoref{figure:mixed}, where LLM-generated content are blended with human-written texts at a four-to-one ratio. 
This mixed dataset challenges the model to differentiate between human and machine-generated content even when combined, making the detection task more difficult. 
The results, shown in the bar and line charts, present a clear comparison between original AUROC and mixed AUROC, as well as the relative performance drop across the four datasets: News, HumanEval, Code, and Yelp Review.

The mixed AUROC scores highlight how each model performs in the face of hybrid content, and the relative drop percentages reveal how the models' performances degrade under this challenging scenario. 
Notably, \ourmethod continues to outperform other models in this more complex task. 
While models like RAIDAR and Binoculars suffer substantial performance drops, especially on Yelp Review and News, \ourmethod maintains stable performance across all datasets. 
For instance, in Yelp Review, where other models experience a significant decrease in AUROC, \ourmethod experiences a modest decrease of just 4.75\%, which is substantially smaller than the declines observed in competing models.
This indicates that \ourmethod is well-equipped to handle mixed content, detecting machine-generated text with greater accuracy even when it is interspersed with human-written segments. 
The drop percentage line, clearly showing the relative drop in performance, emphasizes the robustness of \ourmethod, as its performance decline remains relatively contained when compared to other methods.

Furthermore, in cases where other models experience dramatic degradation (e.g., R-Detect and Ghostbuster on Yelp Review), \ourmethod not only shows higher robustness but also performs consistently across various domains. 
This reinforces \ourmethod’s ability to generalize to different types of content. 
This is a key strength of the method, as real-world applications often involve dealing with content that is not purely human- or machine-generated but a mixture of both. 
The model’s ability to handle this mixed content without a dramatic performance drop demonstrates that its stylistic divergence framework is both robust and adaptable to diverse real-world scenarios.
These findings directly address RQ2, showing that \ourmethod excels not only in clean, unperturbed datasets but also under adversarial conditions and when handling mixed datasets.
The model offers significant advantages over existing baselines in both robustness and reliability.
Its ability to maintain high detection performance despite the complexity introduced by adversarial attacks and data mixing further validates its practical applicability in dynamic environments where such challenges are prevalent.

\subsection{Explainability and Pluggability Analysis}
\subsubsection{Visualizing Explainability through UMAP}
To highlight the explainability of our proposed method, \ourmethod, we utilize UMAP~\cite{mcinnes2020umapuniformmanifoldapproximation} to visualize the distribution of feature vectors extracted from various datasets. The results are shown in \autoref{figure:explainability-UMAP}. UMAP provides a two-dimensional projection that facilitates the analysis of high-dimensional data, making it easier to understand the separability of human-written and machine-generated text. The visualizations confirm the efficacy of \ourmethod in distinguishing between these two types of content across multiple datasets and conditions, thereby validating its explainability.

The top row presents the UMAP projections for the original datasets, which include News, HumanEval, Code, and Yelp Review. In each of these projections, the red points represent machine-generated text, while the green points correspond to human-written text. A striking feature across all the datasets is the clear separation between the two groups, with the red and green points clustered in distinct regions. Notably, the Essay dataset (shown in the rightmost subfigure) exhibits a particularly clean and pronounced separation. This observation visually confirms that \ourmethod is capable of capturing style-based differences between human and machine-generated content, even when the content is complex or nuanced, such as in the Essay dataset.

The bottom row of the figure shows the UMAP projections after the datasets have been subjected to adversarial perturbations. These perturbations are designed to test the robustness of \ourmethod in the presence of noise and adversarial changes. While the News and Code datasets (depicted in the second and third subfigures of the bottom row) show some overlap between clusters, with perturbations causing a slight mix of human and machine-generated points, the majority of the points still remain distinguishable. This is a key finding, as it demonstrates that \ourmethod retains its ability to separate human and machine-generated texts, even when the data has been modified to introduce adversarial noise. The clear separation seen in the Yelp Review and HumanEval datasets further emphasizes the resilience of \ourmethod in distinguishing text styles.
Interestingly, the degree of overlap in the perturbed News and Code datasets indicates that \ourmethod may encounter some challenges in these domains when exposed to adversarial perturbations. 
However, the fact that a majority of the data points are still distinguishable suggests that \ourmethod's underlying feature space is robust and can still identify key stylistic differences, even under challenging conditions. 
This reinforces the strength of the approach and highlights the explainability and reliability of the model in real-world applications where adversarial inputs are common.

\subsubsection{Quantitative Explainability through Metrics}
\begin{figure}[!t]
    \centering
    \footnotesize
    \setlength{\tabcolsep}{2.0pt}
    \includegraphics[width=\linewidth]{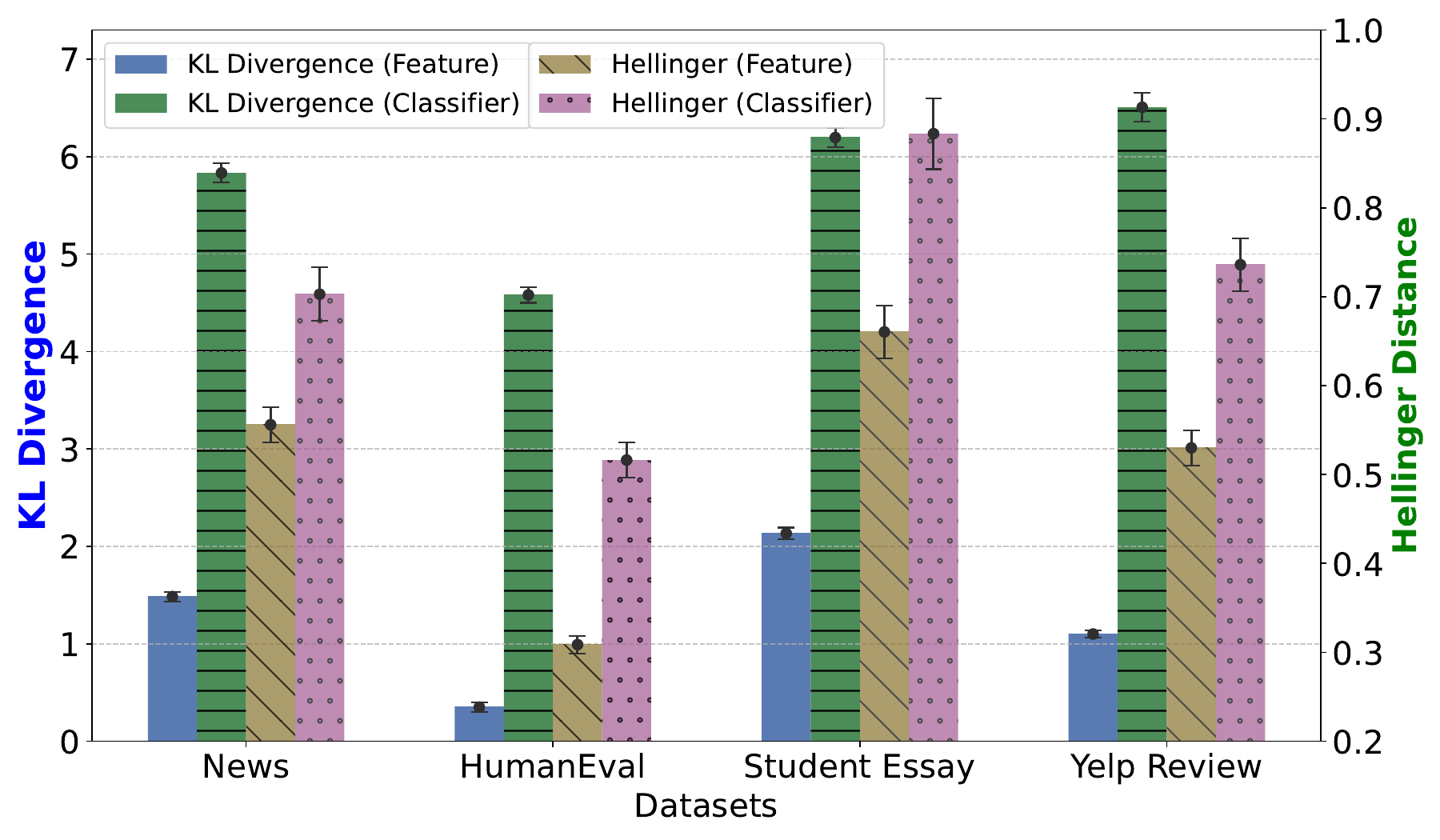}
    \caption{Explainability experiments using KL Divergence and Hellinger Distance for feature and classifier-based methods across multiple datasets}
    \label{figure:explainability-metric}
\end{figure}
While the UMAP visualization in the previous section provided a qualitative view of the feature space separation, we quantify the model's ability to distinguish between human and machine content using two distribution divergence metrics: KL Divergence and Hellinger Distance.
These metrics measure the divergence between the distributions of human-written and machine-generated texts, providing a deeper understanding of how the model distinguishes between the two. 
The results are presented in \autoref{figure:explainability-metric}, where two methods are used: the first derives distributions from the feature $v(x)$ extracted  by \ourmethod, and the second uses classifier prediction scores $\sigma(\hat y)$.
The results consistently show that the classifier-based method (using $\sigma(\hat y)$) outperforms the feature-based method (using $v(x)$) in both KL Divergence and Hellinger Distance across all datasets. 
In particular, the KL Divergence is significantly higher for the distribution $\sigma (\hat y)$, especially in domains like Yelp Review, suggesting that the classifier captures more distinct stylistic differences. 
Similarly, higher Hellinger Distance values for the distribution $\sigma (\hat y)$ indicate clearer separability between human-written and machine-generated texts.
These findings demonstrate that \ourmethod not only achieves high detection accuracy but also enhances transparency and explainability. 
By using these metrics, \ourmethod provides clear, explainable signals that improve trust and usability, making it a reliable tool for distinguishing between human and machine-generated content in practical applications.

\subsubsection{Pluggable Performance across Diverse Text Domains}
This experiment evaluates the pluggability of different text representation methods across diverse domains, including news articles, programming code, student essays, and Yelp reviews. 
The aim is to assess how well each method captures semantic meaning within these various text types and to understand whether some models perform better in specific domains. 
As shown in~\autoref{table:pluggability}, the results reveal the versatility of our approach, where different models can be integrated seamlessly into our \ourmethod framework, without significant degradation in performance.

% -------------
\begin{figure}[!t]
    \centering
    \footnotesize
    \setlength{\tabcolsep}{2.0pt}
    \includegraphics[width=\linewidth]{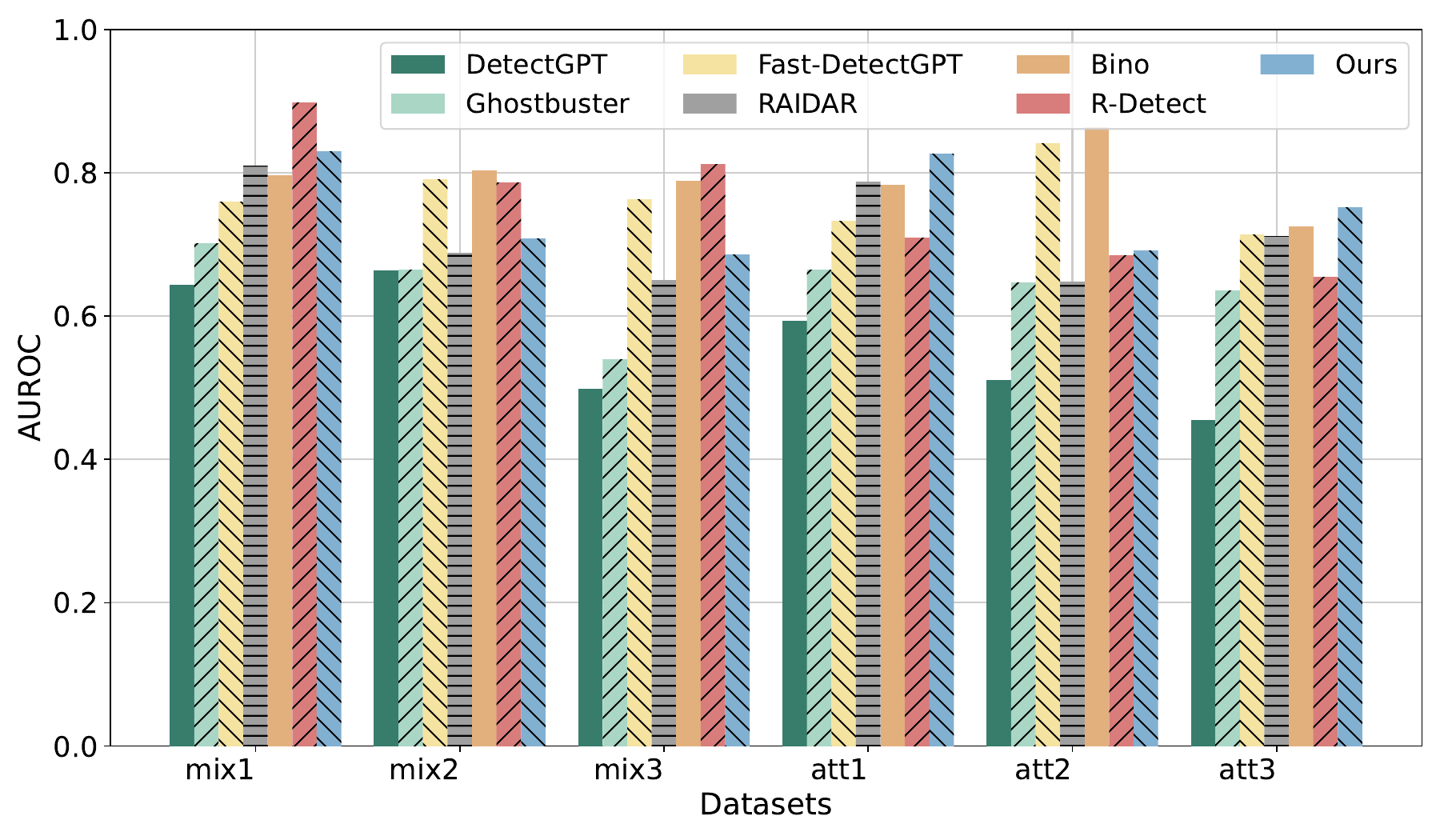}
    \caption{AUROC comparison of LLM-generated text detection on RAID datasets across multiple detection models}
    \label{figure:RAID}
\end{figure}
% -------------------

\begin{table}[!t]
  \centering
  \footnotesize
  \setlength{\tabcolsep}{2.0pt}
  \caption{Performance of various text representations for the pluggability experiment across different domains.}
  \label{table:pluggability}
  \begin{tabular}{lcccc}
    \toprule
    \textbf{Method} & \textbf{News} & \textbf{HumanEval} & \textbf{Student Essay} & \textbf{Yelp Review} \\
    \midrule
    TF-IDF & 0.8683 & 0.7709 & 0.9285  & 0.8114 \\
    Word2Vec/GloVe & 0.8864 & 0.8087 &0.9369 & 0.8309 \\
    BERT & 0.9300 & 0.7812 & \textbf{0.9478} & 0.8566 \\
    SBERT & 0.9356 & \textbf{0.8108} &  0.9455& \textbf{0.8718}\\
    \bottomrule
  \end{tabular}
\end{table}
We tested four prominent text representation techniques: TF-IDF, Word2Vec/GloVe, BERT, and SBERT. While simpler models like TF-IDF provide reasonable results in certain cases, they consistently underperform compared to more advanced models, such as Word2Vec/GloVe, BERT, and SBERT. These advanced models notably improve performance, especially in complex and context-dependent domains like news articles and student essays. Among them, BERT shows superior performance in domains requiring nuanced understanding, like news and academic writing, demonstrating its capability to capture deep contextual relationships. On the other hand, SBERT excels in handling specialized domains like programming code and user-generated content, such as Yelp reviews, due to its focus on sentence-level embeddings and semantic alignment.
This experiment illustrates that \ourmethod, by incorporating pluggable models, offers substantial flexibility, allowing for the integration of different embedding techniques depending on the task at hand. Whether the task involves technical language, nuanced academic texts, or user feedback, \ourmethod can be adapted to maximize performance by selecting the most suitable text representation method. 
This highlights the adaptability and effectiveness of \ourmethod in diverse real-world applications, further validating its practical utility and scalability.

\subsection{RAID Dataset Detection Performance}
To further validate the effectiveness of our method, we conducted experiments on the RAID dataset~\cite{dugan2024raid}, a recent and more challenging benchmark for detecting machine-generated text. 
The dataset includes a range of configurations to assess the detection performance under varying conditions. 
We followed the experimental setup of R-Detect, using a data shuffling mechanism to divide the data into multiple configurations: three original mixtures of machine-generated texts and three corresponding datasets subjected to random attacks. 
These configurations feature texts generated by several state-of-the-art language models, including \textit{GPT-3}, \textit{GPT-4}, and \textit{MPT-30B}, to ensure diverse text sources.

The results indicate that \ourmethod performs very competitively across all configurations. 
While the method does not always achieve the highest detection scores, especially when compared to some of the other state-of-the-art methods, it consistently holds its ground, particularly in configurations where the texts have been subjected to random attacks. 
In these perturbed datasets, \ourmethod outperforms many of the competing approaches, demonstrating its robustness even in the presence of adversarial manipulation. 
The method delivers strong performance in configurations with mixed machine-generated texts, maintaining a higher AUROC score than several well-established methods, which suggests that it is highly effective at distinguishing between human and machine-generated content, even when the data is not pristine.

Moreover, the results show that \ourmethod holds up well in more complex settings involving various attack strategies. Although some of the baseline methods outperform it in specific configurations, particularly with clean data, \ourmethod stands out in scenarios where random perturbations are applied. This resilience under challenging conditions highlights the model's generalizability and its capability to maintain performance across a wide range of real-world scenarios where data integrity may be compromised. Overall, the results firmly establish \ourmethod as a robust and reliable method for detecting machine-generated text, with a performance level closely comparable to or exceeding that of the current state-of-the-art methods in certain configurations.

\section{Conclusion}
In conclusion, \ourmethod provides an effective and adaptable solution for detecting machine-generated text across various domains. 
By leveraging stylistic divergence, it achieves high detection accuracy and robustness, even under adversarial perturbations and data mixing. 
The method’s flexibility allows it to seamlessly adapt to different text types and models, ensuring strong performance in both in-domain and cross-domain scenarios.
Additionally, its explainability and transparent attribution mechanism make \ourmethod a reliable tool for real-world applications. 
Overall, \ourmethod outperforms existing methods in detection reliability and adaptability, marking a significant advancement in machine-generated text detection.

\bibliographystyle{IEEEtran}
\bibliography{ref}

\end{document}